\documentclass{article}
\usepackage{INTERSPEECH2018,amsmath,graphicx,multirow,array}
\usepackage{url}
\newcolumntype{P}[1]{>{\centering\arraybackslash}p{#1}}
\DeclareMathOperator*{\argmax}{argmax}

\title{Improved ASR for Under-Resourced Languages Through Multi-Task Learning with Acoustic Landmarks}
\name{Di He$^1$, Boon Pang Lim$^2$, Xuesong Yang$^3$, Mark Hasegawa-Johnson$^3$, Deming Chen$^1$}
\address{
  $^1$Coordinated Science Lab, University of Illinois at Urbana-Champaign, Urbana, Illinois, USA 61801\\
  $^2$Novumind Inc, Santa Clara, USA 95054\\
  $^3$Beckman Institute, University of Illinois at Urbana-Champaign, Urbana, Illinois, USA 61801}
\email{dihe2@illinois.edu, bplim@novumind.com, xyang45@illinois.edu, jhasegaw@illinois.edu, dchen@illinois.edu}
\begin{document}
%
\maketitle
\begin{abstract}
Furui first demonstrated that the identity of both consonant and vowel can be perceived from the C-V transition; later, Stevens proposed that acoustic landmarks are the primary cues for speech perception, and that steady-state regions are secondary or supplemental.  Acoustic landmarks are perceptually salient, even in a language one doesn't speak, and it has been demonstrated that non-speakers of the language can identify features such as the primary articulator of the landmark.  These factors suggest a strategy for developing language-independent automatic speech recognition: landmarks can potentially be learned once from a suitably labeled corpus and rapidly applied to many other languages. This paper proposes enhancing the cross-lingual portability of a neural network by using landmarks as the secondary task in multi-task learning (MTL). The network is trained in a well-resourced source language with both phone and landmark labels (English), then adapted to an under-resourced target language with only word labels (Iban).  Landmark-tasked MTL reduces source-language phone error rate by 2.9\% relative, and reduces target-language word error rate by 1.9\%-5.9\% depending on the amount of target-language training data.  These results suggest that landmark-tasked MTL causes the DNN to learn hidden-node features that are useful for cross-lingual adaptation.
\end{abstract}
\section{Introduction}
\label{sec:intro}

In the early 1980s, Furui~\cite{Furui1983} demonstrated that the identity of both consonant and vowel can be perceived from a 100ms segment of audio extracted from the C-V transition; in 1985, Stevens~\cite{Stevens85} proposed that acoustic landmarks are the primary cues for speech perception, and that steady-state regions are secondary or supplemental.  Acoustic landmarks produce enhanced response patterns on the mammalian auditory nerve~\cite{Delgutte1984}, and it has been demonstrated that non-speakers of a language can identify features such as the primary articulator of the landmark~\cite{Jyothi2015}.
Automatic speech recognition (ASR) systems have been proposed that depend completely on landmarks, with no regard for the steady-state regions of the speech signal~\cite{Juneja2003}, and such systems have
been demonstrated to be competitive with phone-based ASR under 
certain circumstances.  Other studies have proposed training two separate sets of classifiers, one trained to recognize landmarks, another trained to recognize steady-state phone segments, and fusing the two for improved accuracy~\cite{hasegawa2005landmark} or for reduced
computational complexity~\cite{dihe2jasa}.  It has been difficult to
build cross-lingual ASR from such systems, however, because very few
of the world's languages possess large corpora with the correct 
timing of consonant release and consonant closure landmarks manually
coded. In this paper we propose a different strategy: we propose to use
reference landmark labels in only one language (the source language).  A landmark detector trained in the source language is ported to the target language in two ways: (1) by automatically detecting landmark locations in target language test data, and (2) by using landmark detection as a secondary task for the purpose of training a triphone state recognizer that can be more effectively ported cross-lingually.  The neural network is trained with triphone state recognition as its primary task; landmarks are introduced as a secondary task, using the framework of  multi-task learning (MTL)~\cite{Caruana1998}.
%
%
%

\begin{figure}[htbp]
  \centering
    \includegraphics[width=8.5cm]{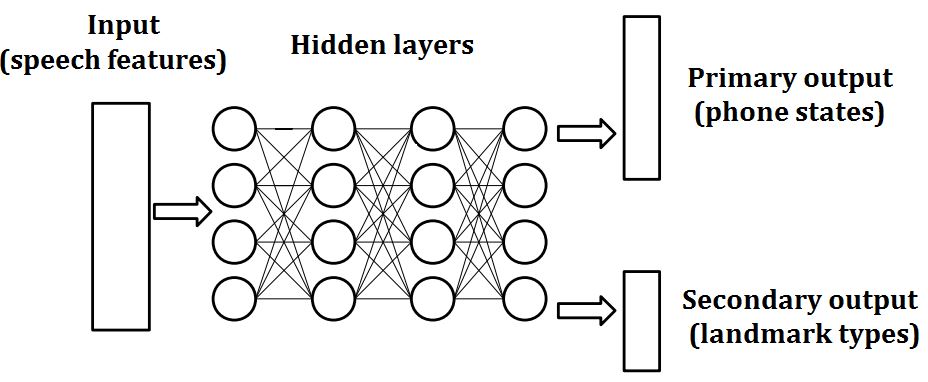}
    \caption{MTL Neural Network Jointly Trained on Phone States and Landmark Types}
    \label{fig:mtl}
\end{figure}

MTL has shown the ability to improve the performance of speech models, especially those based on neural networks~\cite{7178814,6639012,6854673,yang2018joint}.  MTL is a mechanism for reducing generalization error.  A single-task neural net is provably optimal, for large enough training datasets: as the size of the training dataset goes to infinity, if the number of hidden nodes is set equal to the square root of the number of training samples, the difference between the network error rate and the Bayes error rate goes to zero~\cite{Barron1994}.  MTL is useful when the training dataset is too small to permit zero-error learning~\cite{7178814}, or when the training dataset and the test dataset are drawn from slightly different probability distributions (e.g., different languages). In either case, MTL proposes training the network to perform two tasks simultaneously.  The secondary task is not important during test time, but if the network is forced to perform the secondary task during training, it will sometimes learn network weights (and consequently, hidden layer activation functions) that are either (1) less prone to over-fitting on the training data than a single-task network, or (2) generalize better from the distribution of the training data to the distribution of the test data.  Landmark detection could potentially be an ideal secondary task for automatic speech recognition (ASR; Fig~\ref{fig:mtl}), since it detects instantaneous events that are informative to phone recognition. Because landmarks have been demonstrated to correlate with non-linguistic perceptual signals (e.g., enhanced response on the auditory nerve~\cite{Delgutte1984}) and because features of a landmark can be classified by non-speakers of the language~\cite{Jyothi2015}, it is possible that the secondary task of landmark detection and classification will force a neural net to learn weights that are more useful for cross-language ASR adaptation~\cite{xiang2016landmark} than those of a single-task network. These characteristics are especially helpful for under-resourced languages: in an under-resourced language, training data may be limited, e.g., there may be little or even no transcribed speech. A Landmark-based system trained on a well-resourced language might be adapted to an under-resourced language, thus improving ASR accuracy in the under-resourced language.
%
%
%
%
%
%

The key contributions of this work are experimental findings supporting the hypothesis that landmark-task MTL reduces the word error rate of a cross-lingually ported ASR.
After we review some background in Sec.~\ref{sec:back}, key methodology and techniques used to apply the Landmark theory to MTL are explained in Sec.~\ref{sec:method}. Results are presented in Sec.~\ref{sec:result}, and the paper concludes in Sec.~\ref{sec:deci}.

\section{Background}
\label{sec:back}

Before we talk about our methodology, we would like to briefly review MTL as a neural network training method and talk about the under-resource corpus we used in this study.

\subsection{Multi-task Learning}
\label{methods:mtl}

Multi-task Learning (MTL)~\cite{Caruana1998} has shown the ability to improve statistical model performance by jointly training a single model for multiple purposes. The 
multiple tasks in MTL share the same input, but generate multiple outputs predicting likelihoods for a primary and one or more secondary tasks. When the multiple tasks are related but not identical, or (in the ideal case) complementary to each other, MTL models offer better generalization from training to test corpus~\cite{7178814}. A number of works~\cite{7178814,6639012,6854673} have proved MTL to be effective on speech processing tasks. Among them~\cite{6854673} proved MTL effective at improving model performance for under-resourced ASR.



When we conduct MTL, for the same input $x$, we prepare two sets of labels. The label $l_{i}^{ph}$ specifies the phone or triphone state associated with a frame, while $l_{j}^{la}$ encodes the presence and type of acoustic landmark. The network is trained in order to minimize, on the training data, a multi-task error metric as shown in Eq.~\ref{eq:eq3}, where $P^{ph}_i(x)$ ($1\le i\le C^{ph}$) is the probability of monophone or triphone state $i$ at frame $x$ as estimated by the neural network, $P^{la}_j(x)$ ($1\le j\le C^{la}$) is the probability of landmark label $j$ at frame $x$ as estimated by the newtork, and $\alpha$ is a trade-off value we use to weight the two sets of labels. We sweep through a small list of candidate $\alpha$'s to find the value that returns the best result on development test data.
\begin{equation}\label{eq:eq3}
\mathcal{L}_{mtl} =(1-\alpha) \sum_{i=1}^{C^{ph}}( l_{i}^{ph}\log(P^{ph}_i(x))) + \alpha \sum_{j=1}^{C^{la}} (l_{j}^{la}\log(P^{la}_j(x)))
\end{equation}


\subsection{The Iban Corpus}
\label{methods:iban}

The under-resourced language studied in this paper is Iban~\cite{samson:hal-01170493}. Iban is a language spoken in Borneo, Sarawak (Malaysia), Kalimantan and Brunei. 
The Malay phone set is similar to English, e.g., the two languages have the same inventory of stop consonants and affricates;
Malay also has a relatively transparent orthography, in the sense
that the pronunciation of a word is usually well predicted by its 
written form.  If Iban orthography is as transparent as Malay, and if its phone set is as similar to English, then it is possible that a landmark detector trained on English may perform well in Malay.  Iban is also selected for these experiments because of the recent release of an Iban training and test corpus with particularly good quality control~\cite{samson:hal-01170493}.  The Iban corpus contains 8 hours of clean speech from 23 speakers.  Seventeen speakers contributed $6.8h$ of training data, and the test-set contains $1.18h$ of data from 6 speakers. The language model was trained on a $2M$-word Iban news dataset using SRILM~\cite{stolcke2002srilm}.
%
%
%
%
%
\section{Methods}
\label{sec:method}

We trained an ASR on the TIMIT corpus using the methods of multi-task learning (Sec.~\ref{methods:mtl}), using the detection and classification of landmarks (Sec.~\ref{sec:markLand}) as a secondary task.  The same ASR is then adapted cross-lingually to the Iban corpus (Sec.~\ref{methods:iban})

\subsection{Defining and Marking Landmarks}
\label{sec:markLand}
Landmark definitions in this paper, listed in Table~\ref{tab:landmark_rules}, are based primarily on those of~\cite{liu1996landmark}, with small modifications. Modifications include the elimination of the 
+33\% and -20\% offsets after the beginning or before the end of some phones, reported in~\cite{liu1996landmark} and~\cite{hasegawa2000time}, in favor of the simpler definitions in Table~\ref{tab:landmark_rules}.
%
%
  \begin{table}[htbp]
    \vspace{-2mm}
  	\caption{\label{tab:landmark_rules} { Landmark types and their positions for acoustic segments, where `c', and `r' denote consonant closure, and release; `start', `middle', and `end' denote three positions across acoustic segments, respectively.}}
  	\vspace{2mm}
  	\centering
  	\begin{tabular}{|c|c|}
  		\hline
  		Manner of Articulation & Landmark Type and Position \\
  		\hline
  		Vowel & V: middle\\
  		Glide & G: middle \\
  		Fricative & Fc: start, Fr: end\\
  		Affricate & Sr,Fc: start, Fr: end\\ 
  		Nasal & Nc: start, Nr: end \\
  		Stop Closure & Sc: start, Sr: end \\
  		\hline
  	\end{tabular}
  \end{table}
  
\begin{figure}[htbp]
    \includegraphics[width=8.5cm]{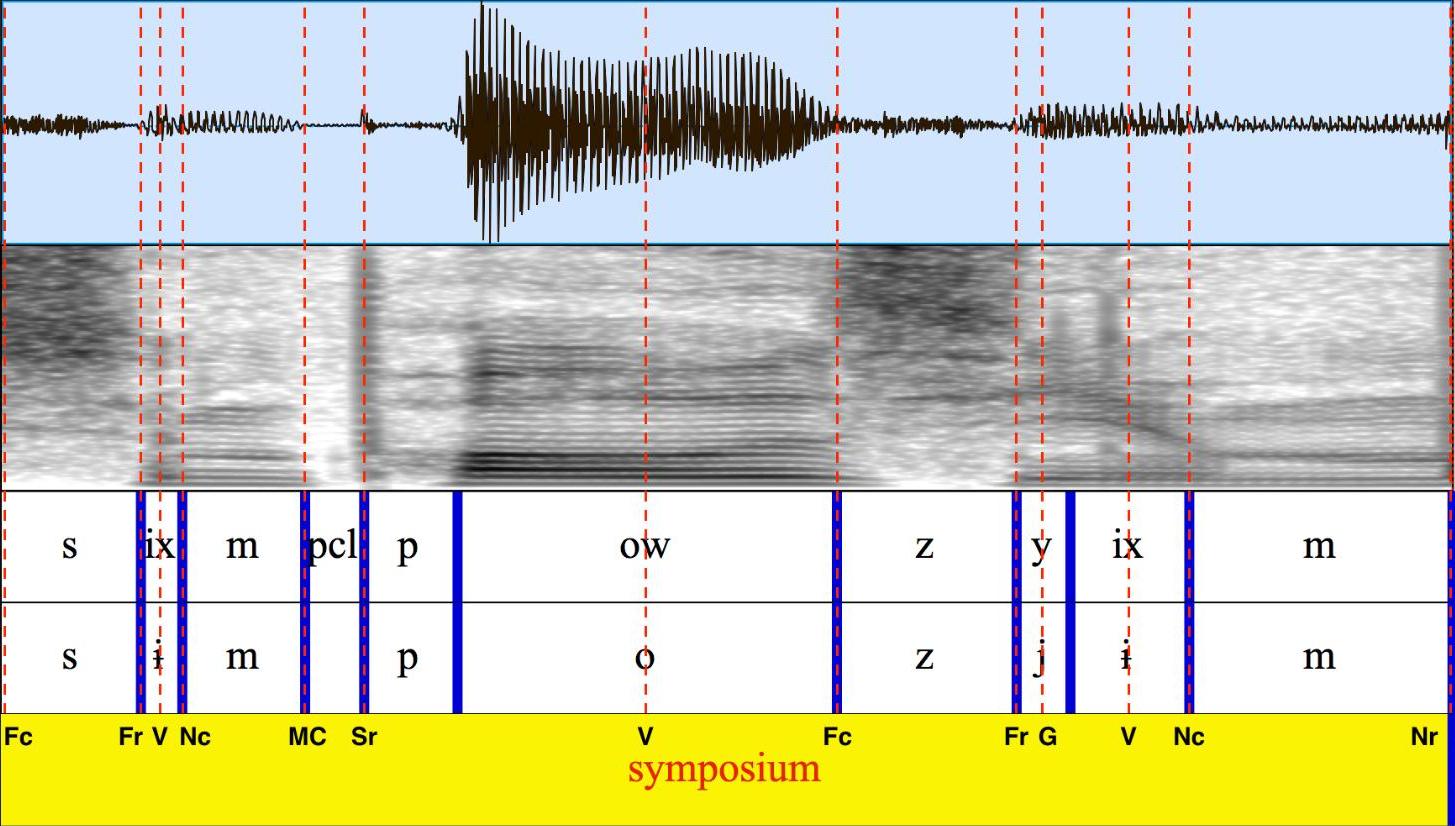}
    \caption{Acoustic landmark labels for the pronunciation of word ``Symposium''.}
    \label{fig:landmark_eg}
\end{figure}

  We extracted landmark training labels by referencing the TIMIT human annotated phone boundaries. An example of the labeling is presented in Fig~\ref{fig:landmark_eg}. This example from~\cite{dihe2jasa} illustrates the labeling of the word ``Symposium''\footnote{selected from audio file: \url{TIMIT/TRAIN/DR1/FSMA0/SX361.WAV}}. The figure is generated using Praat~\cite{boersma1996praat}.
  
  Landmarks are relatively infrequent compared to phone-state-labeled speech frames: every frame has a phone label, but fewer than 20\% of frames have a Landmark label. Because of the sparsity of Landmark-labeled frames, we explored different ways to adjust the Landmark labels to achieve the best MTL performance. We found, expanding the range of a Landmark to include the nearby 2 frames returns the highest accuracy for the primary task.
  
  To further address the imbalance among different Landmark classes, the training criterion was computed using a weighted sum of training data, with weights inversely proportion to the class support.

\subsection{Cascading the MTL to Iban}\label{sec:casc_iban} 
\begin{figure*}[htbp]
    \includegraphics[width=17cm, height=5cm]{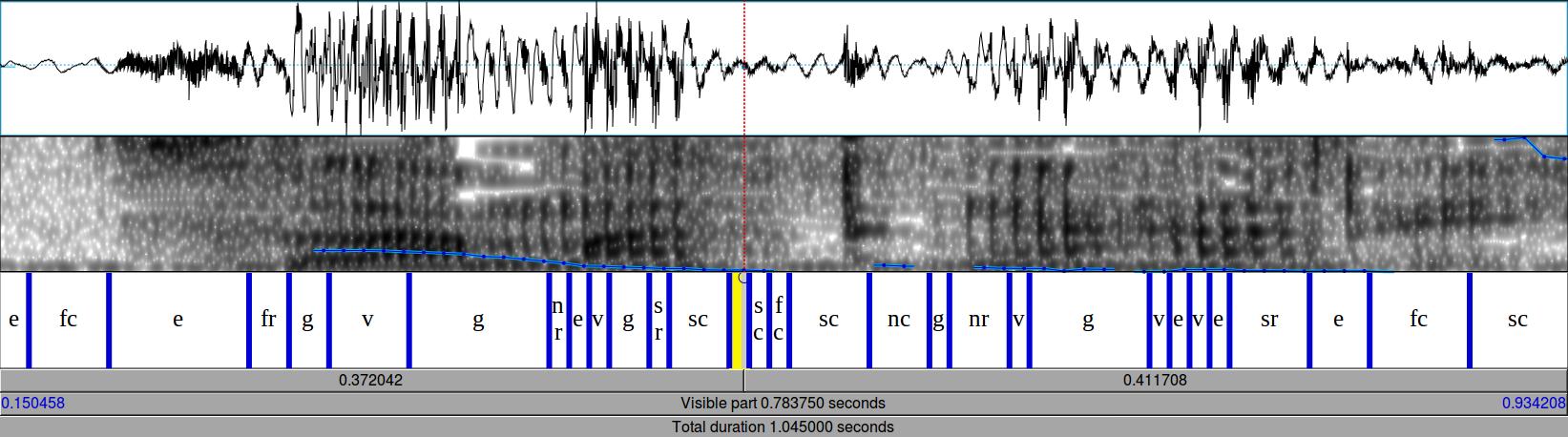}
    \caption{Landmark Detection Result on Iban for utterance ibm\_003\_049, pronouncing {\bf selamat tengah ari} ({\bf s-aa-l-a-m-a-t t-aa-ng-a-h a-r-i} in Iban phone set).  Transcription labels: e=empty (no Landmark); fr, fc, sr, sc, nr, nc, v, g are as in Table 1.}
    \label{fig:iban_landd}
\end{figure*}

After we trained a landmark detector on TIMIT, we ran the detector on Iban. The English-trained landmark detector output is used to define reference labels for the secondary task of the Iban acoustic model MTL. An example of the detector output on an arbitrary utterance\footnote{\url{iban/data/wav/ibm/003/ibm_003_049.wav}} in Iban is given in Fig~\ref{fig:iban_landd}. We found that the results are good at outlining fricative landmarks. The detector can also find stop closure landmarks near the correct locations, but with less precision than the fricative landmarks. The performance on vowel and glide landmarks is only fair: the detector often mixes up the two classes, and incorrectly labels sonorant consonants as vowels.

When applying the landmark detector to Iban, we are concerned with the error generated by the detector. The automatically detected landmark labels are treated as ground truth for MTL in landmark-task MTL in Iban, therefore it is possible that erroneously detected landmarks may mis-lead the network training. To minimize the effect of these mistakes, we introduce an extra weighting factor in the MTL training criterion based on the confidence of the landmark detector output, as shown in Eq.~\ref{eq:eq4}.
\begin{equation}\label{eq:eq4}
\mathcal{L}_{x} =(1-\alpha c_x) \sum_{i=1}^{C^{ph}}( l_{i}^{ph}\log(P^{ph}_i(x))) + \alpha c_x \sum_{j=1}^{C^{la}} (l_{j}^{la}\log(P^{la}_j(x)))
\end{equation}
where $c_x$ is a confidence value derived based on the landmark detector output for feature frame $x$ based on Eq~\ref{eq:eq5}.
\begin{equation}\label{eq:eq5}
c_x=P^{la\_de}_m(x)-\frac{1}{C^{la}-1}\sum_{k=1,k\ne m}^{C^{la}}(P^{la\_de}_k(x))
\end{equation}
where $P^{la\_de}_i(x)$ is the softmax output for landmark class $i$. The class index $m=\argmax_i^{C^{la}} P_i^{la\_de}(x)$, which is also the index for the class the landmark detector predicted.  

The intuition behind this extra layer of weighting is to assign a penalty, during training of the ASR, that is proportional to our certainty of its error. If the detector is not confident separating the output class from other classes, then we reduce the loss it generates in the MTL process.

We experimented with multiple ways to initialize the landmark detector and the phone recognizer in the second language. We found that using a network trained through MTL in TIMIT to initialize the MTL network in the second language yields the best results. We found the technique marginally but consistently outperforms other initializations including DBN.

\section{Results}
\label{sec:result}

All experiments were conducting using the Kaldi~\cite{povey2011kaldi} toolbox.
We extracted an acoustic feature vector using the same algorithm and parameters as~\cite{6639012}. The acoustic model (AM) is a deep neural network with 4 hidden, fully-connected layers, 2048 nodes/layer. The same features and network structure were used for both the landmark detector, the MTL model and the baseline. The baseline is initialized using a DBN~\cite{mohamed2012acoustic}. No speaker adaptation is used in any of the ASR systems in this paper.

Results are reported in Table~\ref{tab:err_rate} for both English (TIMIT) and Iban. TIMIT results are reported to indicate the performance of Landmark-based MTL in the source language, prior to cross-language adaptation.

On development test sets in both corpora, the value $\alpha=0.2$ returned the lowest error rate (with little variability in the range $0.1\le\alpha\le 0.3$), and was therefore used for evaluation. The landmark detector achieves 80.11\% frame-wise accuracy in validation. Phone error rate (PER) was reasonably good: 20.6\% for the baseline system, and 20.0\% for the MTL system, as compared to 22.7\% for the open-source Kaldi tri4\_nnet recipe. 

Decoding results for Iban are reported using Word Error Rate (WER), because the Iban corpus is distributed with automatic but not manual phonetic transcriptions. The comparison between PER in TIMIT and WER in Iban permits us to demonstrate that Landmark-based MTL can benefit PER in a source language (English), and WER in an adaptation target language (Iban).  Triphone-based ASR trained without MTL on TIMIT, then adapted to Iban, achieves 18.4\% WER; a system that is identical but for the addition of landmark-task MTL can achieve 17.93\% WER.  Neither system includes speaker adaptation, and therefore neither system is better than the 17.45\% 
state of the art WER for this corpus\footnote{\url{https://github.com/kaldi-asr/kaldi/blob/master/egs/iban/s5/RESULTS}} with the same language model.

  \begin{table}[htbp]
  	\caption{\label{tab:err_rate} {\it Decoding Error Rate for mono-phone (Mono) and tri-phone (Tri) on TIMIT and Iban.}}
    \vspace{2mm}
    \centering
\begin{tabular}{|P{1.3cm}|P{0.8cm}|P{1.4cm}P{1cm}P{1.5cm}|}
\hline
Corpus                                                                      & AM   & \multicolumn{1}{l|}{Baseline} & \multicolumn{1}{l|}{MTL} & MTL w/ Confid \\ \hline
\multirow{2}{*}{\begin{tabular}[c]{@{}l@{}}TIMIT \\ (PER)\end{tabular}}     & Mono & 24.6                          & 24.2                     & NA            \\ \cline{2-2}
                                                                            & Tri  & 20.6                          & 20.0                     & NA            \\ \hline
\multirow{2}{*}{\begin{tabular}[c]{@{}l@{}}Iban-full \\ (WER)\end{tabular}} & Mono & 24.62                         & 24.22                    & 24.18              \\ \cline{2-2}
                                                                            & Tri  & 18.40                         & 18.03                    & 17.93              \\ \hline
\multirow{2}{*}{\begin{tabular}[c]{@{}l@{}}Iban-25\%\\ (WER)\end{tabular}}  & Mono & 28.87                         & 27.97                    & 27.64              \\ \cline{2-2}
                                                                            & Tri  & 21.31                         & 20.70                    & 20.63              \\ \hline
\multirow{2}{*}{\begin{tabular}[c]{@{}l@{}}Iban-10\%\\ (WER)\end{tabular}}  & Mono & 31.16                         & 28.49                    & 28.48              \\ \cline{2-2}
                                                                            & Tri  & 25.12                         & 23.64                    & 23.57             \\ \hline
\end{tabular}
\end{table}

As we can see in Table~\ref{tab:err_rate}, in all cases, regardless of AM and corpus, the ASR system jointly trained with landmark and phone information returns lower error rate. The setups "Iban-25\%" and "Iban-10\%" train the AM on only 25\% (100 minutes) and 10\% (40 minutes) of the training data uniformly selected at random from the Iban training set (maintaining speaker and gender ratio), but evaluates the error rate on the full test set. As the amount of training data decreases, the benefits of MTL increase. When only 10\% of training data is available, simulating a very low resource case, MTL reduces the word error rate by the greatest margin: 8.7\% for monophone ASR and 6.17\% for triphone ASR. Weighting the MTL loss according to confidence results in a small but consistent error rate reduction. All systems use the same language model, and all systems use acoustic models with the same network architecture and feature set; the error rate change we observe is caused entirely by the use of landmark-task MTL.

\section{Discussion and Future Work}\label{sec:deci}
This demonstrates that landmark-task MTL results in a neural network that can be more effectively ported cross-lingually. As the amount of training data in the under-resourced language is reduced (from 400 minutes to 100 or 40 minutes), the benefits of landmark-task MTL increase.  In addition, introducing a loss weighting according the landmark detector confidence seems to reduce the effect of landmark detector error as it consistently produces lower error rate.  

While a cross-language Landmark detector provides useful information complementary to the
orthographic transcription, visual inspection indicates that a cross-language landmark detector is
not as accurate as a same-language landmark detector. Future work, therefore, will train
a more accurate landmark detector, using recurrent neural network methods that do not depend on human-annotated phone boundaries, and that can therefore be more readily applied to multi-lingual training 
corpora.

\section{Acknowledgements}
This research was partially supported by the Qatar National Research Fund (QNRF) grant 7-766-1-140.
%
%

\bibliographystyle{IEEEbib}
\bibliography{strings,refs}

\end{document}